# Learning Robust Video Synchronization without Annotations

Patrick Wieschollek*[†], Ido Freeman* and Hendrik P.A. Lensch*
*University of Tübingen
[†]Max Planck Institut for Intelligent Systems

*Abstract*—Aligning video sequences is a fundamental yet still unsolved component for a broad range of applications in computer graphics and vision. Most classical image processing methods cannot be directly applied to related video problems due to the high amount of underlying data and their limit to small changes in appearance. We present a scalable and robust method for computing a non-linear temporal video alignment. The approach autonomously manages its training data for learning a meaningful representation in an iterative procedure each time increasing its own knowledge. It leverages on the nature of the videos themselves to remove the need for manually created labels. While previous alignment methods similarly consider weather conditions, season and illumination, our approach is able to align videos from data recorded months apart.

## I. Introduction

Understanding and assessing the current situation around us in a single glance is necessary to interact with the world as we know it. Looking at Figure 1 within few tens of milliseconds, we can verify that the highlighted frames show the same scene although they are recorded at different times, having different illumination, motion blur and seasonal effects. For this task, we are focusing with ease on useful traits like the location of the houses or the shape of the street and are masking irrelevant features such as traffic or the actual road condition. From our experience we are familiar with all effects that might change the current appearance and understand the global context.

For computing a temporal *video alignment* (synchronization), *i.e.* a dense frame-to-frame mapping for each time step in the involved videos with consistency along the time dimension, classical approaches are based on matching local descriptors [1], [2], [3], [4]. While they allow for accurate video alignments, they are limited to video pairs of short video clips sharing very similar appearance. However, longer videos capturing the same content at different times might look completely different besides additional challenges such as ego- and object-motion or changes of the view angle and illumination.

Though data-driven approaches like deep convolutional neural networks have proven excellent performance and capabilities of global scene understanding [5], they usually require a large amount of high-quality labeled training data. One way to automate the labeling process would be to record synchronization signals such as Longitudinal Time Codes, genlock, GPS data or a landmark-based audio fingerprinting [6] during acquisition. Solutions based on this kind of additional data are as accurate as the device which registered the data. Besides its limitation to out-door scenes, GPS has a common precision of three meters [7]. While minor alignment errors would not be visible in vanilla image alignment, any such non-frame accurate alignment would become apparent during simultaneous video playback.

In many real-world scenarios, these explicit synchronization signals are not available as most consumer cameras only encode the creation date of the video file within the meta-data. Directly applying learning based approaches like [8] to learn the alignment is not possible in this case. Producing a dense labeling by manual effort is not feasible either[1]. Our dataset consists of 28 million frames. Note, the ILSVRC challenge [9] is based on 1.4 million *labeled* images only. Particularly, in our setting, we deal with unstructured video content *without* any explicit knowledge about which frames or entire videos do match or not.

To overcome these problems, we propose a novel learning-based approach:

– Section III introduces a new challenging dataset for video-alignment covering rural scenes as well as city scenes across a year under different appearances.
– In Section IV we propose a novel training protocol for training a neural network to match frames from different videos of the same scene *without* any annotation.
– Section IV-C presents a novel method for robust identification and computation of matching tours for partially overlapping video pairs, which is able to automatically detect start and end points of the matching tour.

## II. Related Work

The process of video alignment holds a natural relation to image alignment which was addressed by several studies [10], *e.g.* using stereo correspondence estimation [11] or robust pixel descriptors [12]. Algorithms like video stitching for creating panoramic videos [2], automatic summarization of videos [13], HDR video generation [14], vehicle detection for advanced driver assistance systems [15] and video-copy detection [16] among others are heavily dependent on such a robust and accurate temporal alignment of video-frames between multiple videos.

Basic video alignment is commonly used in the field of human action retrieval or surveillance motion capture. Here, finding similarities of human actions in videos are based on dynamic time warping of various sensor features to track the human skeleton [17]. Bazin *et al*. introduced ActionSnapping [18] which focuses on synchronizing actions performed by humans such as weightlifting, baseball pitching or dancing assuming a static background scene and frontal views.

---

[1]Thoroughly aligning a video pair of 8 min length by hand takes 41min.

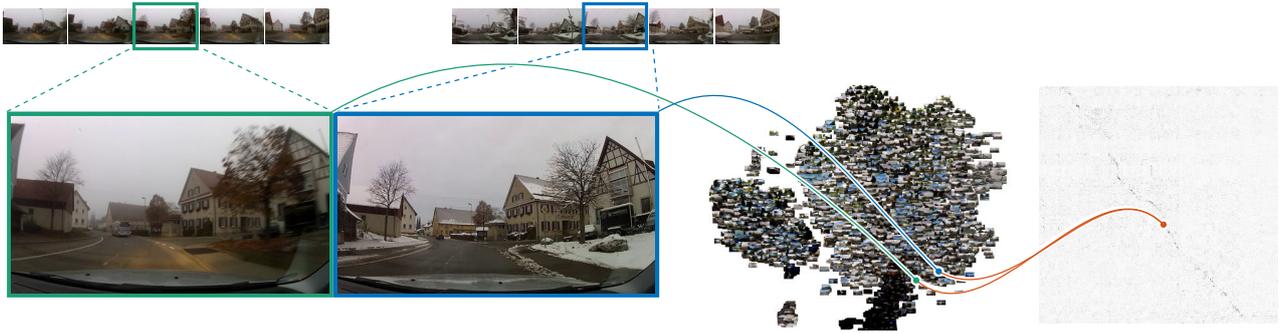

Fig. 1. Given a pair of videos (top) our network has learned to robustly embed each frame (left) in a high dimensional feature space (middle) independently from appearance changes caused by weather, traffic or seasonal effects. Given these feature vectors, the algorithm retrieves the most plausible frame mapping for a synchronous playback (dark path, right). Our approach autonomously learned to compare these frames during training *without* any human annotation or labels.

Most approaches for spatio-temporal video alignment [19], [20], [21], [22], [23] assume a linear temporal correspondence, *i.e.* a constant time shift between successive frames or a constant change in playback speed for one video. Sand and Teller [3] compute a matching-likelihood of 3D motion to match videos. A non-linear solution was proposed by Wang *et al*. [4] based on a matching histogram of SIFT features using nearest neighbour search. Here, the search space increases with the length of the video sequence. In terms of application the most similar work to ours is from Evangelidis *et al*. [24], [25], which allows for sub-frame accurate alignments of at most one minute video snippets under negligible appearance changes on rather simple street scenes. Another related task is to recognize places from different view angles. A data-driven version of VLAD descriptors by Arandjelovic *et al*. [8] demonstrates the capability of neural networks to detect specific locations. Compared to place recognition video alignment however requires a much finer resolution. Figure 2 illustrates typical examples of frame-pairs from different datasets which are considered as similar for the specific task.

Learning similarities by training neural networks was done previously for very specific applications such as signature verification [26], face recognition [27], [28] and comparing image patches for depth estimation [29]. They rely on datasets with extensive human annotations and reliable ground-truth data.

## III. DATASET

Our underlying dataset comprises of 602 full-HD, 30FPS videos (1.8 TB of raw data) capturing 260 hours of commuters' car journeys on partially overlapping routes between April 2012 and March 2013 and spanning over approximately 16,000km. The videos were captured using a GoPro Hero 2 camera mounted on the dashboard before every journey without special adjustments. However, the view angle does not feature differences as large as in place recognition tasks [8].

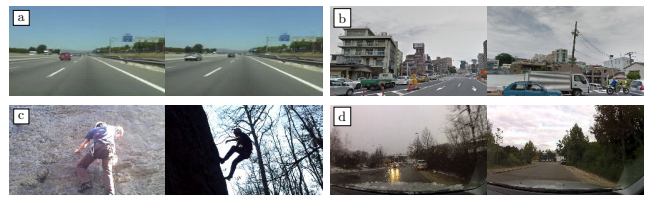

Fig. 2. Comparing the data sets of related work. Each pair shows a typical matching frame-pair. a) [24] only shows small changes in appearance and perspective, b) [30] has a totally different view angles and c) [31] requires only a constant time shift for alignment. For our dataset d), an alignment method has to handle a different appearance and produce a playback with non-linear speed.

The acquired videos feature both rural landscapes and urban scenes under varying traffic conditions such as temporary road works, rush hour, diverse weather conditions, *e.g.*, snowfall, rain, and seasonal environment appearance, for example effects of vegetation as well as different daytime illumination (see Figure 2d) and the supplemental video https://youtu.be/vhVsw4qoe70).

Most videos show journeys between the same two cities but still have a variation in start and end locations and the actual roads driven. The temporal alignment of the videos thus poses a further challenge when trying to match stationary situations like waiting in traffic during rush hour with a video showing little traffic. This is different to linear video alignment tasks like copy-detection described in [31].

The enormous range of possible variations in the video content which also suffers from a noisy acquisition (*e.g.* different viewing angles, wipers, raindrops, camera transformation, etc.) requires some understanding of context of the entire scene rather than a simple local feature matching like histogram based methods [4], [18], [25]. Moreover, the videos might show interrupted content, *e.g.* when the lens is cleaned or remounted during recording.

## IV. UNSUPERVISED CURRICULUM LEARNING

We propose a framework to learn descriptors for all frames such that the Euclidean distance represents a similarity metric between captured scenes or locations. By properly designing the training protocol this generic approach features fast matching computation, robustness against seasonal effects and it does not rely on pre-existing labels for training.

The similarity metric can be exploited to robustly synchronize videos as will be explained in Section IV-C. While first establishing the procedure to align partially overlapping video pairs $(X, Y)$ it is straightforward to extend it to multiple videos in a collection (see Figure 13).

*a) Curriculum Learning.:* The algorithm alternates between two steps: the *learning* step and the *label generation* step. In the learning step, we assume given labels $\ell$ and train a neural network to produce meaningful descriptors for a similarity metric $\delta$. The label-generation is based on the current version of the trained network combined with tour matching (Section IV-C) to exploit temporal consistency. Given the learned similarity metric $\delta$ novel, potentially more reliable labels are produced, replacing the old ones.

Over multiple iterations, more and more sophisticated and more informative training data is generated. None of these steps requires any human annotation nor recorded GPS signals.

### A. Learning Step

In the learning step we want to learn an embedding of frames to establish a similarity measure between individual frames $(x, y)$. Relying on the currently available labels from the training data we train a convolutional neural network (CNN) [32] to predict a high-dimensional descriptor $\Phi_i$ for each frame $i$ such that the Euclidean distance

$$(x, y) \mapsto \delta(x, y) := \|\Phi_x - \Phi_y\|_2 \quad (1)$$

is small when the frames are similar and vice versa. We use the standard ResNet-50 architecture [33] and add a projection from the *pool5* layer to learn the 1000-dimensional descriptors $\Phi_i$. As CNNs tend to learn edge filters in the first layers, we use a pre-trained ResNet version for object recognition as initialization.

In order to efficiently train the concept of similarity the triplet neural network approach [34] with weight-sharing is used, which generalizes well to unseen examples. It requires labels $\ell = (a, p, n)$ in the form of triplets of frames: for an anchor frame $a$ the label needs one similar or positive frame $p$ and one negative/dissimilar frame $n$. The similarity metric $\delta$ (Eq. (1)) is enforced by minimizing the triplet loss by Hoffer *et al.* [34]

$$L(a, p, n) = \left[m + \|\Phi_a - \Phi_p\|_2^2 - \|\Phi_a - \Phi_n\|_2^2\right]_+ \quad (2)$$

for some margin $m \in \mathbb{R}$. The first term penalizes embeddings of similar frame-pairs $(a, p)$ that are too far away from each other in the high dimensional feature space. The latter penalizes embeddings of negative (non-matching) frame-pairs $(a, n)$ if they are too close to each other (closer than some margin $m$ – Figure 3). In practice, we constrain this embedding to live on the $d$-dimensional hypersphere, *i.e.* $\|\Phi_i\|_2 = 1$ and set $m$ to $0.5$.

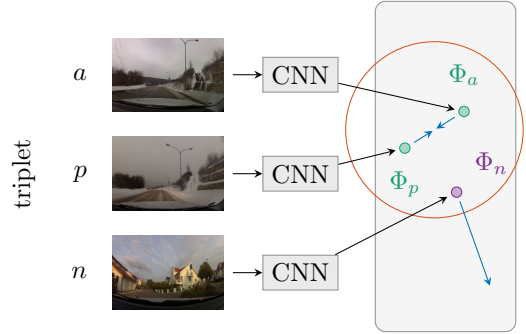

Fig. 3. Let $(a, p, n)$ be a frame-triplet with descriptors $(\Phi_a, \Phi_p, \Phi_n)$ produced by the same CNN. During training, the CNN is optimized to embed positive frames $p$ close to the anchor point $a$, while negative frames $n$ are pulled away from the anchor point if the distance is smaller than a margin $m$.

### B. Label Generating Step

While the learning step is rather straight-forward the challenge lies in automatically generating appropriate frame-triplet labels as training data. This label-generating step *automatically harvests* new training data for subsequent learning steps by explicitly exploiting the coherence in videos and by proposing and judging video alignments based on current knowledge. The goal is to gather more and more informative training data in each iteration by successively increasing the difficulty, *i.e.* to find positive frame pairs which show the same scene location but with potentially different appearance as well as finding negative pairs which currently are assigned rather similar embeddings. This is achieved in three waves:

Iteration 0: *intra-video* sampling of nearby frames for initial training

Iteration 1+: *inter-video* sampling of frames from matching tours

Iteration 2+: *transitive* inter-video sampling of frames from matching tours by propagation of alignments to other videos

After each iteration of the label-generating step we re-trained the neural network in the learning step, alternating between training and label generation.

*a) Iteration 0.:* In Iteration 0 one has to solve the dilemma of generating reliable labels $\ell$ without having any trained network for proposing distances $\delta$. Instead we rely on the inherent coherence within the same video. Any arbitrary frame-pair which is at most 15 frames apart serves as positive sample $(a, p)$. Any other random frame sufficiently far away from $a$ is regarded as a negative frame $n$.

*b) Iteration 1+.:* After the first iteration the network is trained and can now produce features $\Phi_i$ for each frame $i$ which is carried out for every tenth frame of all videos in the data set. One can use the proposed $\Phi_i$ for approximating the pair-wise similarity, but since the network is not fully trained

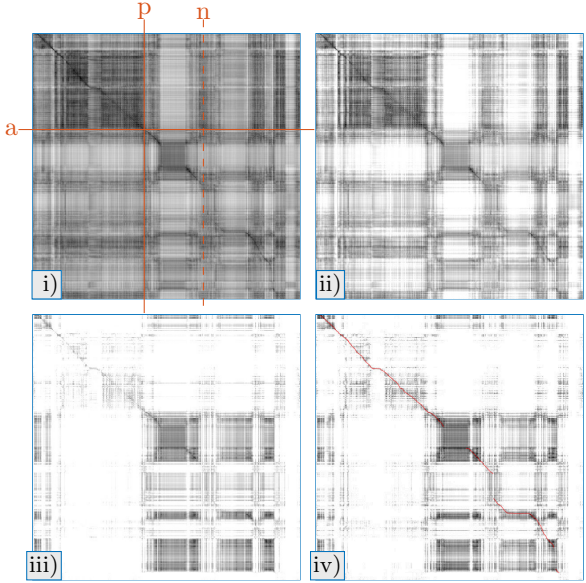

Fig. 4. Development of the cost matrix of the same video pair throughout the iterative training process. Dark entries represent frame-pairs that are considered similar having a small distance in the feature space. Note how the path (red) of similar frame-pairs becomes more distinct. Large coherent regions along the path indicate stand-still, e.g. waiting at a traffic light. We further highlight matching frames $a, p$ and hard negative $n$.

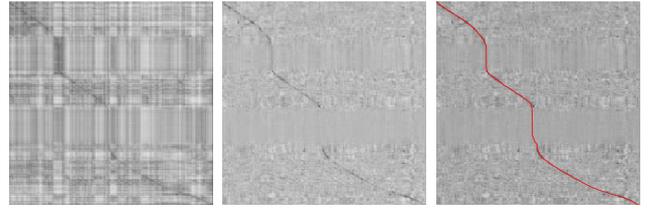

Fig. 5. Computed similarity matrix from the first iteration in our training process (left). As they contain highly correlated entries, directly finding a matching tour would fail. After the de-correlation of the cost matrix (middle) finding a path (right) is significantly more robust.

yet it is essential to judge whether we can rely on a particular frame encoding or not.

The criteria for accepting a positive frame pair or detecting a negative one in this step is based on temporal coherence by potential *matching tours* between *distinct* videos $(X, Y)$. A cost matrix

$$C = (\delta(\Phi_x, \Phi_y))_{x,y}, \quad x \in X, y \in Y \quad (3)$$

can be computed containing all coarse pair-wise distances of frames. Corresponding frames should have a small distance. Though this might not always be the case in early iterations, it might already be possible to detect a matching tour as explained in Section IV-C, *i.e.* to find an optimal path through the cost matrix (see Figure 4). If the path is sufficiently distinct the path will synchronize both videos correctly and all nodes on the path resemble positive pairs with row $a$ and column $p$, independent of the currently proposed measure $\delta(a, p)$. Some new positive pairs might be found this way giving hints on how to optimize the embedding in the next training phase. Similarly, we can harvest challenging negative pairs $(a, n)$ by just choosing a column $n$ sufficiently far away from the path. Most informative will be such a pair if $\delta(a, n)$ is currently rather small, indicating similarity though the frames clearly should be rated distinct.

Though we might not find all possible paths between all videos yet, the resulting labels $\ell(a, p, n)$ will be more informative than in the previous iteration as the appearance between two videos will more likely be different even for matching frames. The process is visualized in Figure 4 where the correct matching path becomes more obvious and easier to detect throughout the iterative training.

*c) Iteration 2+.:* Computing a cost matrix for each single video pair is inefficient. Instead, we propagate detected matching tours transitively to other pairs. Based on the property of "$X$ and $Y$ have a matching tour", denoted as $X \sim Y$, is a equivalence relation (meeting the requirements reflexivity, symmetry and transitivity), we propose a transitive sampling. In fact, our entire dataset can be split into distinct equivalent classes $V/\sim$ under the relation $X \sim V$ when videos $X$ and $V$ share a part of the same tour. Hence for any two videos $X$ and $Y \in V/\sim$ we already know the existence of a matching tour $X \sim V$ and $V \sim Y$. Transitivity also directly gives us a matching tour for $X \sim Y$ if both share some overlap. This allows us to sparsely sample video-pairs and propagate matching frames across videos. Using a tree-based index structure reduces the complexity to $\mathcal{O}(n \log n)$ when synchronizing $n$ videos (see Figure 13). Through transitivity one can establish matching tours which so far could not be detected using the previously trained embeddings. With this iterative process we can quickly generate a huge number of challenging and informative training triplets $\ell(a, p, n)$ even for videos which have been captured months apart.

Starting in iteration 2 we mix the obtained training examples from inter-video sampling (iteration 1+) and transitive inter-video sampling (iteration 2+).

### C. Finding Matching Tours

We will now describe how to find a matching tour given a cost matrix $C$. The entire iterative scheme is based on robustly detecting false-positives from the network prediction and producing complex training data in a reliable way.

*1) Pre-Processing: De-correlate Costs:* Particularly in the early iterations, the similarity matrices $C$ produced by the CNN contain a lot of false predictions, since it is not yet fully adapted to the task. These errors exhibit a low-rank structure, because any frame that is not correctly embedded is likely to corrupt an entire column (or row) of the similarity matrix (left of Figure 5). Additionally, some of the pairs are, indeed rather similar although we would like to treat them as different. For example, many journeys through rural areas with little information but crop fields on both sides of the road appear extremely similar. As a pre-processing to the tour extraction we remove those correlation effects by subtracting a low-rank matrix approximation $C' \leftarrow C - \sum_{k=1}^{r} U_k \Sigma_k V_k^\star$ from singular value decomposition $U\Sigma V^\star$ of $C$. The result is illustrated in Figure 5 (middle) using rank $r = 5$. Abusing notation we further denote the de-correlated cost matrix $C'$ as $C$.

*2) Formulation as a Shortest Path Algorithm:* The only missing step for aligning a video pair is to find a plausible path through its respective de-correlated cost matrix. Intuitively, a path is a collection of consecutive frame-pairs of minimal matching costs. Matching frame-pairs should lie on a clearly distinct path in the cost matrix. A well-studied algorithm to solve a shortest-path problem is *Dijkstra's Algorithm*. For a given start- and endpoint it computes the globally optimal path with minimum cost. We shortly outline the vanilla grid-version when applying to the cost matrix $C$.

*a) Dijkstra's algorithm.:* For detecting paths we only allow for three directions: downwards, rightwards and a diagonal bottom-right step basically preventing reverse playback and assuming non-negative costs, *i.e.* $c_{ij} \geq 0$. Given a start-entry $(s,t)$ and end-point $(v,w)$ with $s \leq v$ and $t \leq w$ the algorithm propagates costs $\tilde{C}$ in a dynamic programming approach with entries

$$\tilde{c}_{s,t} = 0, (\tilde{c})_{i,j} = \min\{\tilde{c}_{i-1,j}, \tilde{c}_{i-1,j-1}, \tilde{c}_{i,j-1}\} \quad (4)$$

$$(c_{\text{direct}})_{i,j} = \arg\min\{\tilde{c}_{i-1,j}, \tilde{c}_{i-1,j-1}, \tilde{c}_{i,j-1}\} \quad (5)$$

and infinite costs $\infty$ for not reachable frames. Following the path backwards encoded in $(c_{\text{direct}})$ from $(v,w)$ gives a path $P$ with lowest costs between $(s,t)$ and $(v,w)$. Figure 5 shows such a path from our augmented version.

*3) Augmented Dijkstra's Algorithm:* Unfortunately, we neither know the start- nor end-point in contrast to the vanilla version used in [4] nor can we guarantee that there is a path through the *entire* cost matrix, *e.g.* consider different sub-tours. For most videos of a large collection one does not even know if two videos match at all. We augment Dijkstra's algorithm by processing subsequences individually with the goal to flexibly handle non-matching regions without corrupting the entire path when searching for the global optimum over the entire matrix.

The coarse cost matrix $C$ is split along one time dimension into multiple overlapping column-stripes $C_0, C_1, \ldots, C_n$ (see Figure 6) each containing 90 seconds of the video. Now, our approach tries to find a matching tour through the entire cost matrix building on possible tours from each stripe.

*a) Local tours within stripes.:* Let us consider such a single stripe $C_k$. Introducing an artificial start node with zero costs to the left enables almost complete freedom regarding the location of each match within one stripe. We are only interested in finding a matching tour from the left $c_{i,0}$ to the right $c_{j,N}$ in the current stripe. Further, this artificial start node allows us to treat each stripe individually

For the final extraction we remove path parts from the overlap –taking all information but the overlap– as Dijkstra tends to deviate (see Figure 7) from the correct path near the borders of the stripe. For each stripe the vanilla regularized

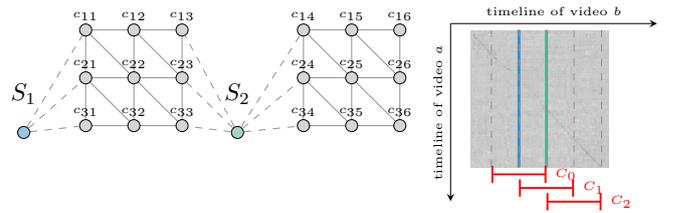

Fig. 6. Given a cost matrix $C$, we split up one video by dividing the cost matrix into overlapping stripes $C_0, C_1, \ldots$. Finding the local shortest path in each stripe independently and testing for plausibility in the overlap region, we obtain a reliable matching tour for synchronizing videos.

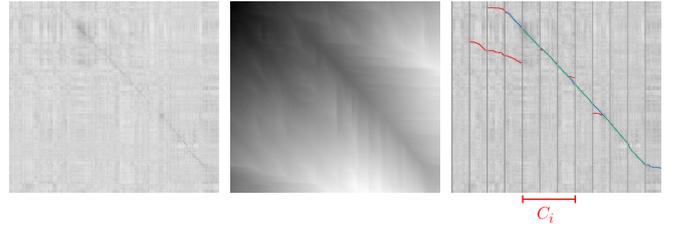

Fig. 7. For any decorrelated cost matrix $C$ (left), we applied our augmented Dijkstra's algorithm unaware of start- and endpoint for overlapping stripes. This gives stripe-wise propagated costs for the entire video snippet (middle). Our heuristic classifies between non-consistent tours-snippets (red, *e.g.* the deviation effect) and reliable tour-snippets (green). The parts illustrated as blue lines will be only taken into account if the do not violate the global consistency constraint. All images are best viewed in the electronic version.

Dijkstra's algorithm is applied as described from $S_t$ to $S_{t+1}$ by only allowing the three mentioned directions. Each possible path $P_t = (S_t, p_1, p_2, \ldots, p_n, S_{t+1})$ has associated *matching costs* defined as

$$\pi(P_k) = \sum_{p \in P_k} c(p), \quad (6)$$

where $c(p)$ is equivalent to an entry $c_{ab}$ in the cost matrix $C$ for the frame-pair $p = (a,b)$.

*Combining local tours to a global matching tour.:* As a robust global matching tour does not necessarily span the entire cost matrix and all stripes, *e.g.* for detours, we reject local stripes which cannot be connected to any neighboring stripes with a tolerance of up to two seconds. This tolerance accounts for the fact that we currently only consider coarse information (each 10-th frame).

Further, applying Dijkstra's algorithm to each stripe individually might generate local paths of minimum costs, which are not necessarily a matching tour – there does not have to be such a matching tour at all. We therefore, reject local stripes with entries $(x,y) \in P_k$ from the local matching tour, if an alternative frame-pair $p' \in \{(x \pm \varepsilon, y), (x, y \pm \varepsilon)\}$ does not have significant higher associated costs than $p$, *i.e.* we simply use the threshold $c(p') > \frac{6}{5}c(p)$. This threshold might be conservative but the specific choice only impacts a small fraction of paths during exploration.

TABLE I
TIMINGS FOR EXTRACTING FEATURES FROM UNSEEN VIDEOS. ENABLING
HIGH FRAME-RATES IS ESSENTIAL ON LARGE-SCALE DATASETS.

| approach | [24], [25] | [4] | [31] | ours |
|---|---|---|---|---|
| fps | 0.11 | 3.77 | 15 | 140 |

All remaining frame-pairs from the global matching tour are considered for the generation of the next training dataset version.

*D. Final Alignment of Videos*

So far, we only considered each 10-th frame during training. In order to prevent visual miss-alignment due to interpolation artifacts in the final alignment the full temporal resolution is required.

A coarse-to-fine approach only computes entries of $\bar{C}$ at finer resolution if they are near a matching tour in the coarse cost matrix $C$. The global matching tour is split into chunks of the same size. For each chunk, we compute all frame distances. As the start and end point of the matching tour through a single box is known, we directly apply vanilla Dijkstra's algorithm without further modification.

Modifying the playback speed of only one video would introduce visible jumps and spurts in this video when matching to the reference video. Consider a linear playback of a reference scene with green traffic light, while the other video has to jump over the frames when waiting on red. To achieve visually pleasing video playback, we smooth the matching tour along both time dimensions using Kalman filters with the additional constraint to not revert the time line of a single video.

## V. RESULTS AND EXPERIMENTS

The following results and timings are obtained on a single workstation with an Nvidia Titan X GPU. We demonstrate the robustness on aligning a couple of challenging scenes. To evaluate our method despite missing ground truth, we compare our convnet prediction of similarity to [24], [25], the SIFT-based histogram matching from [4], conducted a user study and evaluate against a manually alignment of videos.

*A. Timings and memory consumptions*

Our approach compares favorably to a local-feature-based approach concerning the run time (Table I) and has small storage requirements (102 MB for the network weights). When considering every 10-th frame of a single video with length 35 minutes, the embedding takes 45 seconds in total. This allows us to efficiently compute new embeddings of the entire dataset of 260 hours content for the training procedure within less than half an hour using 12 GPUs.

The path detection and extraction procedure to produce coarse paths on unseen video-pairs of 35 minutes each takes two seconds given the embeddings. This splits into pairwise-distance computation (1071 ms, GPU), de-correlation (370 ms, CPU) and path detection/computation (405 ms, CPU). Computing the final matching tour takes 6 seconds due to multiple runs of the path finding procedure on finer scale. This gives a speed-up factor of at least 300 compared to [4], [24], [25]. So far ours is the first approach enabling large-scale interactive applications. See the https://youtu.be/vhVsw4qoe70 for a real-time demonstration. For processing many more videos additional search structures might be used.

*B. What is the network looking for?*

For aligning videos, the network has to distinguish between relevant and irrelevant regions in the frame, *e.g.* the appearance of traffic and road lanes and the weather depend on the moment of recording. To visualize which input information are used inside the neural network for a particular prediction, we compute saliency maps using guided-ReLU [35]. Informally, it computes the gradient information of the network output wrt. to the input images holding all weights fixed. This indicates which pixel information in the input image have large impact on the network prediction. Compared to the vanilla ResNet-50 our trained model learned to ignore irrelevant information like traffic, see Figure 9. Instead it focuses on the shape of the horizon and vegetation of the environment. This is not possible by previous methods [18], [3], [4], [24], [25] as they also put attention on passing cars and clouds as depicted in the lower row of Figure 9.

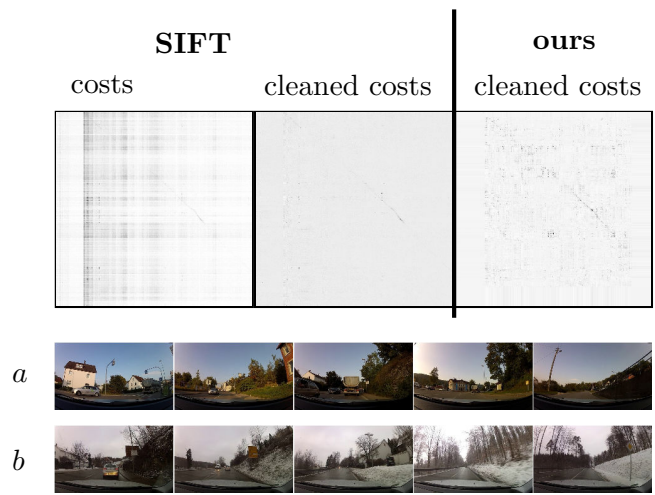

Fig. 11. Although, there should be a matching tour between frames in $a$ and $b$ through the entire video snippet, it is not possible to align both videos using SIFT features [4]. Our network was able to learn similar feature embeddings even between these videos taken in September 2012 and February 2013 respectively. The cost matrix in our approach contains reasonable information for most frames.

*C. Robustness and Accuracy*

In contrast to methods solely based on aggregating SIFT features, our method is able to even match videos captured five months apart as depicted in Figure 11. Remarkably, as the videos for the dataset are collected over multiple months the network has learned to interpret scenes globally. Even sequences where human interventions like tree-felling cause a rather different look of the same scene, the corresponding

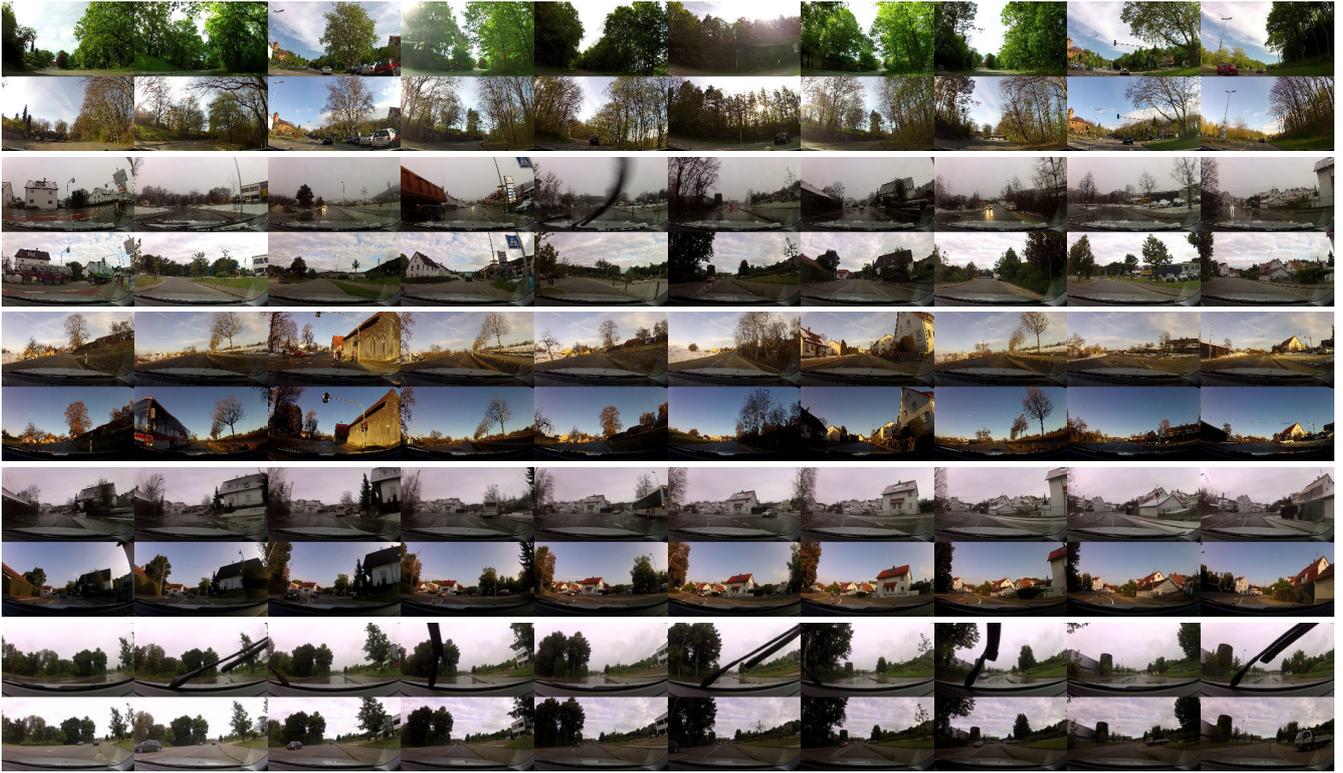

Fig. 8. Each row contains pairs of coarsely aligned videos (every 10-th frame) across different seasons, lightning, weather conditions as well as vegetation. These videos are taken from the validation set and are not used during training. Notable, the algorithm can robustly handle windscreen wipers, motion blur and raindrops on the wind shield.

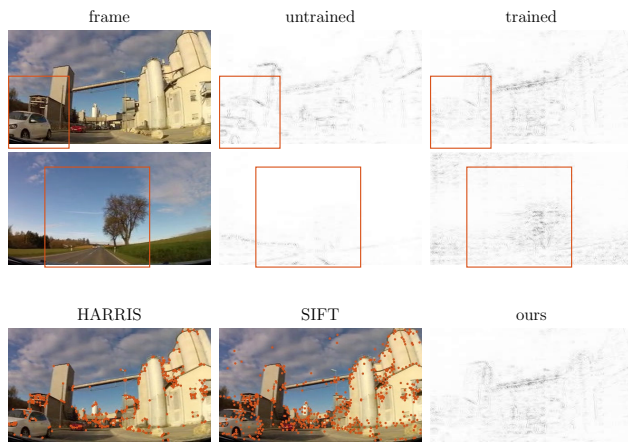

Fig. 9. The network (right) learned to ignore common content such as traffic or road lanes and instead focuses on striking environment information compared to vanilla ResNet (middle). Previous methods [18], [3], [4], [24], [25] rely on local features, capturing irrelevant information like the white car in the lower row.

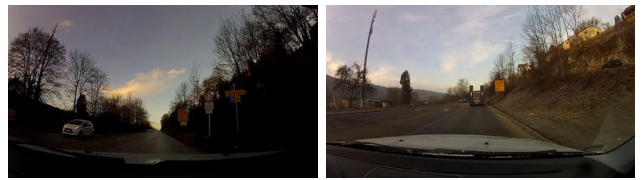

Fig. 10. The left frame was captured in November 2012 and right frame in March 2013 after tree felling works. These kinds of differences cause high matching costs of the found tour in our approach, but the situation is resolved by our matching tour procedure exploiting temporal consistency.

videos are correctly synchronized by our approach as we enforce temporal consistency.

To evaluate the accuracy quantitatively, we manually judged 500 tours predicted by our approach from the cost matrices (see Figure 4, left). Note, how the accuracy increases over the iterations. Hence, the harvested additional training data of higher complexity results in a higher recall of found matching tours.

In addition, we thoroughly annotated the videos from Figure 8 for two experiments by manually fitting the best visual matching frame, well knowing the shown locations and temporal context. In a user-study we showed a single reference frame and the manually aligned frames besides several similar neighboring frames. The evaluation of 450 submitted results from 14 participants is illustrated in Figure 8, which reveals discordance between different participants on the same frame.

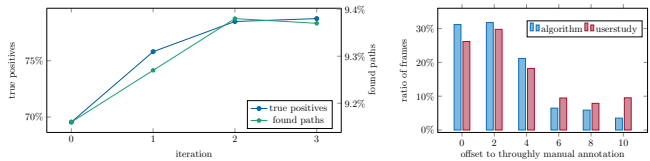

Fig. 12. Comparison to human annotated alignment. After each iteration we report (left) the number of found matching tours between videos and true positives and the mis-alignment (right) against a human annotation.

Approximately, 53% agree with a tolerance of 4 frames. This clearly demonstrates the difficulty of this task, probably caused by changes of camera perspective, lacking temporal information in rural scenes or unfamiliarity with the shown locations. This might also contain effects as in Figure 10. Second, we directly compared the frame-distance between our annotations and the extract path of our approach. This compares favorably to our estimated human performance as 62% of the predicted frame-pairs have a frame offset of at most 4 frames.

*D. Limitations*

Despite its good performance, our approach is subject to a number of limitations. Ignoring the high-res input might drop discriminative small local clues, which could be solved by attention techniques [36], [37]. Spatial Transformer Network Layers [38] which also facilitates optical flow estimation could account for spatial displacement between matching frames caused by the current acquisition method. Recently proposed methods like [39] can be included to also directly learn spatial alignments. While we apply Kalman filtering to smooth the matching tours, one might formulate this as a discrete optimization problem to include the produced matching costs by the neural network.

## VI. CONCLUSION

We present a novel combination of deep neural networks and path finding algorithms for synchronizing videos by approximating the similarity of frame-pairs based on the feature embedding by a deep convolutional neural network. Matching tours between two videos are determined along the resulting frame-based similarity matrices. To improve the extraction of the correct matching tours we propose pre-processing of these cost matrices and a regularized version of Dijkstra's algorithm on cost-stripes to satisfy time constraints. Our training method relies on an iterative scheme to automatically gather new labels completely avoiding manual annotations. The system verifies temporal consistency of predictions to create newly labeled training data. Utilizing the transitivity of matching tours between multiple video-pairs increases the complexity of the input data gradually and allows us to robustly synchronize videos months apart under different weather conditions and vegetation.

The system opens up several exciting directions for future research, *e.g.* label-transfer between densely annotated videos-frames and automatically synchronized videos with different appearances. Hence, the label annotation for large video collections can be done in shorter time and for fewer costs. Learning appearance modification as in [40] or a dynamic version of content blending as described in [41] are further exciting applications, which can benefit from our learning-based approach for video synchronization.

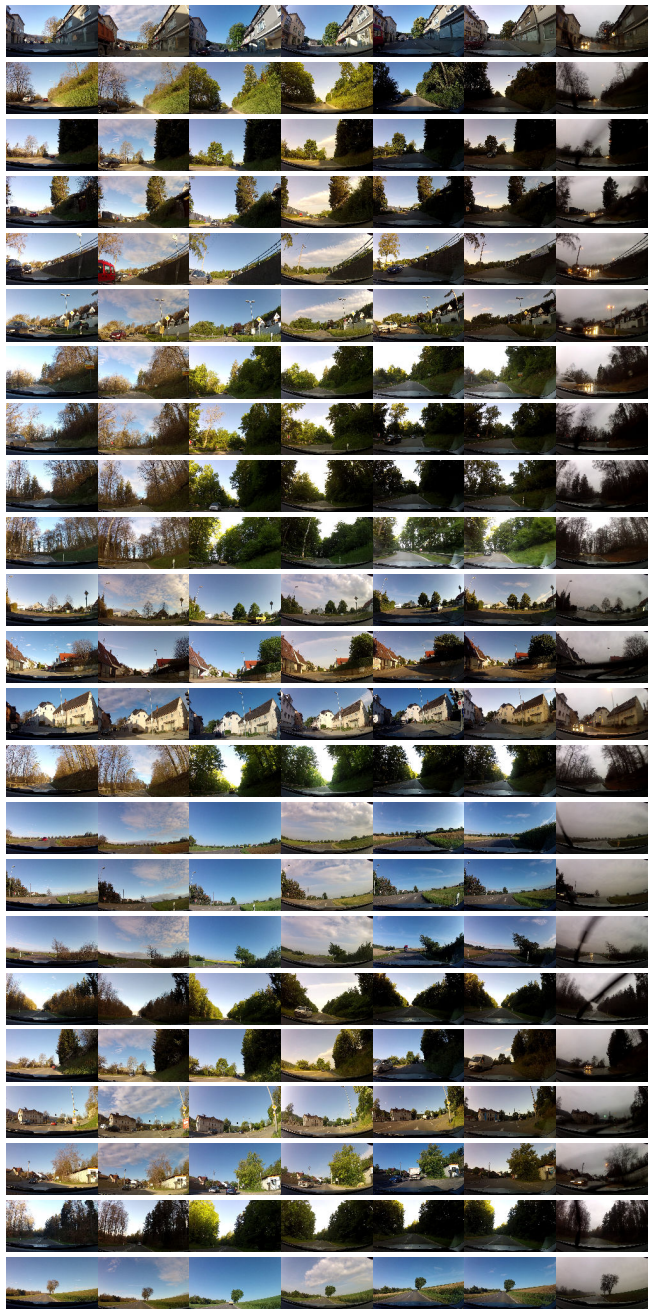

Fig. 13. Multiframe *temporal* alignment: After quering similar and time consistent video snippets using the learned embeddings, the approach is able to robustly synchronize all found snippets to a reference video snippet.